\documentclass[letterpaper, 10 pt, conference]{ieeeconf}  

\usepackage{graphicx}
\usepackage[T1]{fontenc}   
\usepackage{caption}
\usepackage{amsmath}
\usepackage{mathtools}
\usepackage{comment}
\usepackage{subcaption}
\usepackage{float}
\usepackage{mathrsfs} 
\usepackage{balance}
\usepackage{booktabs}
\usepackage{bm}
\usepackage{flushend}
\usepackage{makecell}
\usepackage{tabularx}
\usepackage{multirow}
\usepackage{calc}

\usepackage{colortbl}
\usepackage{xcolor}

\usepackage[detect-none]{siunitx}
\makeatletter
\let\NAT@parse\undefined
\makeatother

\usepackage{url}
\usepackage{hyperref}
\hypersetup{
    colorlinks=true,
    linkcolor={blue},
    citecolor={blue}
}
\usepackage{cleveref}

\usepackage{array}
\usepackage{paralist}
\usepackage{xspace}

\definecolor{Gray}{gray}{0.97}
\definecolor{MedGray}{gray}{0.9}
\definecolor{greytext}{gray}{0.5}
\definecolor{DarkGreen}{rgb}{0.0, 0.5, 0.0}
\definecolor{PFGreen}{rgb}{0.0, 0.5, 0.0}
\definecolor{lightGreen}{rgb}{0.8, 0.9, 0.8}
\definecolor{CadmiumGreen}{rgb}{0.0, 0.42, 0.24}
\definecolor{DarkKhaki}{rgb}{0.74, 0.72, 0.42}
\definecolor{DarkRed}{rgb}{0.7, 0.2, 0.2}
\definecolor{Purple}{rgb}{0.7,0.0,0.7}
\definecolor{Brown}{rgb}{0.7,0.3,0}
\definecolor{Orange}{rgb}{1, 0.5, 0.1}

\newcommand{\hsE}[1]{#1} 

\newcommand{\etal}{\textit{et al.}}
\newcommand{\eg}{e.g., }
\newcommand{\ie}{i.e., }

\graphicspath{
{figures/} 
}

\IEEEoverridecommandlockouts                              

\title{\LARGE \bf
Charting Visual Impression of Robot Hands
}

\author{Hasti Seifi$^{1}$, Steven A. Vasquez$^{2}$, Hyunyoung Kim$^{3}$, and Pooyan Fazli$^{4}$
\thanks{$^{1}$Hasti Seifi is with the School of Computing and Augmented Intelligence, Arizona State University, Tempe, AZ 85281, USA
        {\tt\small hasti.seifi@asu.edu}}%
\thanks{$^{2}$Steven A.\ Vasquez is with the Department of Computer Science,
        San Francisco State University, San Francisco, CA 94132, USA
        {\tt\small svasquez7@sfsu.edu}}%
\thanks{$^{3}$Hyunyoung Kim is with the School of Computer Science, University of Birmingham, Birmingham, B15 2TT, United Kingdom
        {\tt\small h.kim.4@bham.ac.uk}}%
\thanks{$^{4}$Pooyan Fazli is with the School of Arts, Media and Engineering, Arizona State University, Tempe, AZ 85281, USA
        {\tt\small pooyan@asu.edu}}%
}

\begin{document}

\maketitle
\thispagestyle{empty}
\pagestyle{empty}

\begin{abstract}
A wide variety of robotic hands have been designed to date. Yet, we do not know how users perceive these hands and feel about interacting with them. To inform hand design for social robots, we compiled a dataset of 73 robot hands and ran an online study, in which 160 users rated their impressions of the hands using 17 rating scales. Next, we developed 17 regression models that can predict user ratings (e.g., humanlike) from the design features of the hands (e.g., number of fingers). The models have less than a 10-point error in predicting the user ratings on a 0--100 scale. The shape of the fingertips, color scheme, and size of the hands influence the user ratings the most. We present simple guidelines to improve user impression of robot hands and outline remaining questions for future work.

\end{abstract}

\section{INTRODUCTION}

Hundreds of robotic hands have been designed in the last decades. For example, the humanoid robot Pepper has a five-fingered hand with articulated joints~\cite{Pepper}. NAO uses a similar design but with only three fingers and no palm~\cite{NAO}. Other robots, such as the PR2 and Baxter, have metal grippers and/or suction cups~\cite{PR2,Baxter}. A growing number of soft manipulators are designed with novel materials and working principles~\cite{shintake2018soft}. For example, Homberg \etal\ developed a silicone-based pneumatic gripper that can comply with a wide range of object shapes~\cite{homberg2019robust}. New designs appear every year focusing on dexterous manipulation of objects and performance metrics~\cite{billard2019trends,shintake2018soft}. 

Robotic hands are often used in collaborative settings with humans. For example, Baxter and PR2 can work on assembly and manipulation tasks with users~\cite{hamandi2018} 
or engage in social touch (\eg\ hand clapping, hugging)~\cite{fitter2020exercising,block2019softness}. The soft gripper by Homberg \etal\ has been attached to Baxter and trained to grasp a range of household objects~\cite{homberg2019robust}. Socially interactive robots can use hand gestures or direct touch to convey emotion and intent in education, therapy, or service tasks~\cite{johal2016child,burns2021haptic}. 
In these settings, users may merely observe the hand or interact with it through touch. 

\begin{figure}
	\includegraphics[width= 0.49\textwidth]{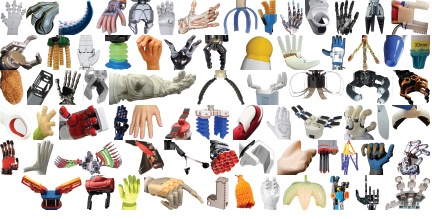}
	\caption{A collage of the 73 existing robotic hands that we evaluated in an online study.}
	\label{fig:teaser}
\end{figure}

However, little is known regarding how people perceive different robotic hands. Despite the vast design space of robotic hands, little effort has been devoted to  studying their design features and how these features affect user impression.
To support the increasing adoption of robots in social settings and guide the design of future robot hands, we investigate how lay users perceive existing robot hands and what design features of the hands (\eg number of fingers) can predict user impression. As a first step, we focus on visual perception of the hands as users normally observe the hands before interacting with them. This initial visual impression can define how the user adjusts their interactions with the robot hand (\eg distance).

We reviewed existing robotic hands and ran an online study to capture lay users' impressions of these hands. 
We first compiled a set of 371 robot hands from commercial and research venues and then selected 73 hands that represented the variations in the set (Figure~\ref{fig:teaser}).
In an online study, 160 users evaluated images of the hands on 17 rating scales (\eg humanlike). 
Next, we created a database of the 73 hands and coded their design features (\eg number of fingers, color scheme).
Finally, we trained 17 linear regression models to predict user ratings from the hand features and analyzed the contribution of the design features to these models.
The shape of the fingertip, color scheme, and hand size were among the top predictors for most ($\geq14$) of the regression models, while the visible surface texture, number of fingers, and existence of a palm only contributed to a few ($\leq3$) predictions. 
We discuss implications of our findings for robot hand design and share our dataset and models to enable future work in this area. 
Our contributions are as follows:
\begin{itemize}
    \item a database of 73 robot hands with 15 design features (\eg number of fingers) and 17 user ratings (\eg humanlike) per hand as well as a comprehensive list of 371 existing hands with their links
    \item 17 regression models that can predict user ratings and an analysis of the models' most predictive design features
\end{itemize}

\section{RELATED WORK}
\label{sec:relatedwork}
We review the literature on robotic hands and present findings on user evaluation of robot appearance. 
In this paper, we use robot ``hand,'' ``end effector,'' and ``manipulator'' interchangeably.

    \subsection{Designing Robot Hands} 
    A wide range of robot hands have been proposed by robotics researchers. One of the earliest robotic hands was a two-fingered parallel jaw gripper that is still in use for many applications.  
    Some designs closely replicated a human hand with articulated fingers and a palm~\cite{controzzi2014humanhand} or were inspired by animals~\cite{kim2013soft}. 
    For example, similar to fish, some robotic end effectors use suction for grasping and moving objects~\cite{Baxter}. 
    Others may closely resemble a tool such as a cup holder or a hook~\cite{Aeolus}. 
    The materials and working mechanisms of the hands have also evolved over the years. 
    Early manipulators 
    were composed of rigid parts and electrical motors~\cite{billard2019trends}. 
    Later efforts have incorporated soft elastic materials (\eg\ electroactive polymers) in designing parts of the hand or the whole hand~\cite{shintake2018soft}. 
    Soft manipulators tend to be smaller due to the actuation limitations of their materials~\cite{billard2019trends}. 
    Past studies have supported and evaluated the wide variety of hand designs according to performance considerations such as weight, speed, ease of design and control, and robustness in interacting with a wide range of objects~\cite{billard2019trends,shintake2018soft}. 
   
    Human-robot interaction (HRI) researchers often use existing robot hands in their studies~\cite{RAS2008,hamandi2018,willemse2016observing,shiomi2017effort}. As an exception, a few recent studies customized the hands of commercial robots to improve user comfort. 
    Fitter \etal\ placed boxing pads over the Baxter's grippers so that users can clap hands with the robot as part of their physical exercise games~\cite{fitter2020exercising}. Zamani \etal\ designed a five-fingered flat hand for the Sawyer robot to study the impact of motion parameters (\eg speed) on user evaluation of a robot-initiated tapping gesture~\cite{zamani2020effects}. The main motivation for designing a custom hand was to measure the applied force, but the authors also noted the addition of a silicone layer for user comfort. 
    These studies relied on the intuition of the researchers in their design and mainly focused on the material softness. We investigate how the materials and other design features of the hand can influence user impressions. 
    
    \subsection{User Impression of Robot Appearance}
    \label{sec:relrobot}
    People can form mental models of a robot and its capabilities based on its appearance~\cite{powers2006robotadvisor,li2010robotappearance}. 
    In a study by Powers and Kiesler, people perceived an anthropomorphic advisor robot with a short chin length to be more sociable and were more likely to follow its advice compared to a robot with a longer chin~\cite{powers2006robotadvisor}. Li \etal\ compared a machinelike robot with an animallike and a humanlike robot in a study and found that the machinelike robot was less likeable than the other two robots~\cite{li2010robotappearance}. 
    
  Recent crowdsourced studies on large collections of social robots have found generalizable trends in user impressions of robots. 
  Reeves \etal\ collected 300 social robots and showed that people evaluate and stereotype robots, similar to their impressions of humans, along two primary dimensions of warmth and competence~\cite{Reeves2020Social}.
  Phillips \etal\ investigated the humanlikeness of 200 robots based on their images~\cite{phillips2018human}. 
   Kalegina \etal\ compiled a database of 176 robots with programmable faces and coded variations in their facial features (\eg the existence of eyelashes)~\cite{kalegina2018}. 
   Based on two studies with 12 and 17 robot faces, Kalegina \etal\ provided guidelines on how different facial features impact user ratings. 
  These studies used images of the robots to collect user impressions on Amazon Mechanical Turk. 
  Similarly, we use Mechanical Turk to collect user impressions of a large set of robot hands. Our focus is on a single body part instead of the whole robot. Therefore, our database lists different types of hands for a robot (\eg vacuum cup gripper and parallel jaw gripper for Baxter~\cite{Baxter}) as separate entries. We also include hand prototypes that are not embedded in existing robots.

  Finally, the literature suggests that users can infer some aspects of a tactile experience through vision. In particular, there is a high correspondence between visual and tactile evaluations of material roughness and hardness~\cite{tiest2007haptic}. The same patterns hold for texture evaluation in the two modalities~\cite{vardar2019fingertip}. Other studies showed that people can also visually infer the affective qualities of materials and vibrations~\cite{fujisaki2015perception,schneider2016hapturk}. In tactile HRI, a recent study found that user pleasantness ratings for videos of a stroking sensation peaked at 3 cm/s, similar to ratings of a physical stroking experience~\cite{willemse2016observing}. 
   Based on these findings, we asked users to judge how they would feel about physical interactions with the hands by looking at the hand images.

\section{Online User Study}
\label{sec:onlinestudy}
To collect the user impressions of robot hands, we compiled a representative set of 73 hands from industry and academia, designed a custom questionnaire, and ran an online user study on Amazon Mechanical Turk. 

\subsection{Compiling Representative Robot Hands}
\label{sec:collect}
    \noindent\textbf{Collecting a Large Set of Hands} -- 
    We use a broad definition for robot hands to capture a large variety of designs. Specifically, our definition includes robotic end effectors that can pick up, hold, or manipulate objects. We also include robot parts that are located at a place that is normally associated with a hand (\eg end of a robot arm) or have the appearance of an animal or human hand. The first part of this definition covers the wide range of robotic grippers~\cite{shintake2018soft} and suction cups~\cite{Baxter}.  
    The second part covers hands with rigid designs such as those in the KASPAR~\cite{huijnen2016matching} or the CuDDler~\cite{6758580} robots. We exclude robotic limbs that are only used for locomotion (\eg~\cite{Aqua}) as well as exoskeletons. 
    Also, we exclude graphical renderings of robot hands or hands that are shown in animations or movies.

    Using this definition, we collected 371 robot hands from existing robot databases and review papers. The authors examined all the robots in the following three databases: (1) IEEE Robots database~\cite{IEEERobots}, (2) Stanford Social Robot Collection~\cite{Reeves2020Social,stanfordcollection}, and (3) ABOT (Anthropomorphic roBOT) database~\cite{phillips2018human,ABOT}. We listed all the robots that had a hand according to our definition. If a robot had multiple hand designs (\eg Baxter~\cite{Baxter}), we added all the designs to our list. This led to 130, 124, and 85 unique hands from these three sources respectively. Finally, we added 32 new designs by examining all the hands presented in recent review papers on robotic manipulators~\cite{billard2019trends,shintake2018soft}. 
    
    \vspace{0.1cm}
    \noindent\textbf{Selecting Representative Designs} --
    We chose 73 robot hands that captured the design variation in the larger set of hands. Four authors separately identified representative hands from the original set and later merged their choices in a meeting. 
    We included all the hands selected by four ($n=10$ hands) or three authors ($n=25$). 
   We discussed the hands selected by one or two authors and reached a consensus on which ones are distinct and should be included in the final set. We added 26 and 12 hands that were selected by two authors and one author respectively.

        \begin{figure}[tb]
         \centering
        \includegraphics[width=0.5\textwidth]{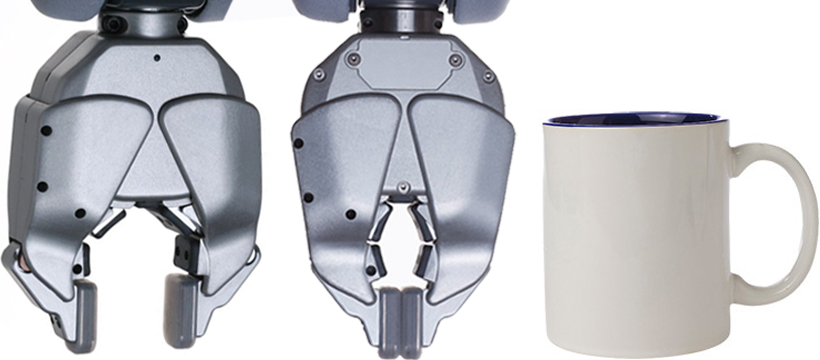}
         \caption{Example of an edited image in the study. The image shows the PR2 gripper and its size w.r.t a medium-sized mug~\cite{PR2}.}
         \label{fig:PR2Scaled}
        \end{figure}

        \vspace{0.1cm}
        \noindent\textbf{Preparing Robot Hand Images} -- 
        We divided the 73 hand models into eight subsets and prepared their images for the study. \hsE{One of the authors grouped the hands with an overall goal of having a variety of hand designs in each of the  sets.} Each set included the edited images of nine unique hands as well as the PR2 gripper for comparison~\cite{PR2}. For each robot hand, we prepared a single image with a white background showing the hand in one or two poses and included a mug or a coin as a scale reference (Figure~\ref{fig:PR2Scaled}). For the majority of the hands, we found one neutral open pose and one closed or figurative pose ($n=49$ out of 73 hands). If one of these two poses was not available online ($n=5$), we included two images from different angles (\eg palmar and dorsal sides of the hand). If the hand was rigid ($n=13$) or if only one pose was available online ($n=6$), we included one image.

    \subsection{Designing a Custom Questionnaire} 

\begin{table}[tb]
\caption{The questionnaire data collected and analyzed in this paper. 
We cite the origin of the questions.}
\label{tab:questionnaire}
\centering
\small{
\begin{tabular}{p{6.5cm}p{1.25cm}}
\hline
\cellcolor{MedGray}\textbf{Question} &  \cellcolor{MedGray}\textbf{Reference}\\
\hline
\textbf{This robot hand is ...}&\\
(1) Machinelike - Humanlike & \cite{bartneck2009measurement,kalegina2018}\\
(2) Creepy - Nice& \cite{IEEERobots} \\
(3) Boring - Interesting& - \\
(4) Incapable - Capable& \cite{carpinella2017robotic} \\
(5) Dangerous - Safe& \cite{carpinella2017robotic} \\
\hline
\textbf{A robot with this hand is ...}&\\
(6) Masculine - Feminine& \cite{kalegina2018}\\
(7) Childlike - Mature& \cite{kalegina2018}\\
(8) Unfriendly - Friendly& \cite{bartneck2009measurement,kalegina2018}\\
(9) Unintelligent - Intelligent& \cite{bartneck2009measurement,kalegina2018}\\
(10) Untrustworthy - Trustworthy& \cite{kalegina2018}\\
\hline
\textbf{If this robot touches me, I would feel ...}&\\
(11) Unhappy - Happy& \cite{bradley1994SAM}\\
(12) Calm - Excited & \cite{bradley1994SAM}\\
(13) Submissive - Dominant& \cite{bradley1994SAM}\\
\hline
\textbf{I feel [Uncomfortable - Comfortable] to ...}&\\
(14) Touch this robot hand.& -\\
(15) Be touched by this robot hand.& -\\
(16) Pass or receive objects from this robot hand.& -\\
(17) If this robot hand interacts with objects near me.& -\\
\hline
\end{tabular}
}

\end{table}
         \begin{table*}[t]
\caption{Design features of robot hands, their range of values, and definitions in our dataset.} 
\label{tab:handfeatures}
\centering
\small{
\begin{tabularx}{\textwidth}{llX}
\hline
\cellcolor{MedGray}\textbf{Hand Feature} &  \cellcolor{MedGray}\textbf{Value} & \cellcolor{MedGray}\textbf{Note}\\  
\hline

Shape of Fingertip &	Pointy, Round, Square, Other, N/A & N/A if there are no fingers in the hand.\\
\hline
Color Schemes & \makecell[l]{Cool, Warm, Mixed (cool + warm),\\ Black, White, Gray, Brown} & List of all noticeable color schemes in the hand.\\
\hline
Size &	Baby, Kid, Adult, Large & Sizes scaled w.r.t a medium-sized mug: $0<$ baby $<0.75$, $0.75\leq$ kid $<1.5$, $1.5\leq$ adult $< 2.25$, and large $\geq2.25+$.\\
\hline
Mechanics Visible &	Yes, No & Yes, if wires, motors, tendons, pipes, etc., are visible.\\
\hline
Has a Thumb &	Yes, No & Thumb should be apart from the other fingers and opposable to them.\\
\hline
Number of Segments in a Finger & 0, 1, 2, \dots & The maximum number of segments in a finger. A finger has 2+ segments if it has a movable joint. All fingers have at least one segment.\\
\hline
Commercial Product & Yes, No & Yes, if the hand is a commercial product.\\
\hline
Even Finger Spacing & Yes, No, N/A & Yes, if the spacing is even between fingers. N/A if the hand has less than three fingers. Human hand has uneven finger spacing.\\
\hline
Rigid &	Yes, No & Yes, if parts of the hand cannot move.\\
\hline
Material &	Metal, Plastic, Rubber, Other & List of all visible materials in the hand.\\
\hline
Has a Palm &	Yes, No & Palm is the area between fingers and the wrist. To have a palm, the hand must have at least one finger, and the finger(s) must be attached parallel to the palm.\\
\hline
Material Softness &	Soft, Hard, Mixed (soft and hard) & Mixed if both types of materials are visible.\\
\hline
Number of Fingers & 0, 1, 2, 3, 4, 5, \dots & Fingers are the terminal members of the hand that resemble or function like a human finger (\eg grasping)\\
\hline
Texture	& Yes, No & Yes, if individual segments of the hand have a visible texture.\\
\hline
Multicolor & Yes, No & Yes, if there are multiple colors in the hand.\\
\hline
\end{tabularx}

}
\end{table*}

         Since no established questionnaire exists for a robotic limb, we designed a custom questionnaire based on past studies of robot appearance and touch interaction. We also used a set of demographic questions from prior work. 
         
           \vspace{0.1cm}
        \noindent\textbf{Robot Hand Questionnaire} -- We aim to capture user impressions of a hand or a robot with this hand as well as their anticipated emotions and comfort in interacting with the hand. 
        Our questionnaire has 17 semantic differential ratings on a 0--100 scale.
        Ten of the ratings are about qualities of the hand (\eg humanlike) or a robot with this hand (\eg intelligent). Eight out of the ten ratings are from the Godspeed questionnaire~\cite{bartneck2009measurement}, the Robotic Social Attributes Scale (RoSAS)~\cite{carpinella2017robotic}, and a recent study on user perception of robot faces~\cite{kalegina2018}. 
        \hsE{While a core principle of the Godspeed and RoSAS questionnaires is that of increasing internal reliability, using a large number of ratings from these questionnaires would lead to increased study fatigue in the participants. Thus, we employ a subset of items from these questionnaires to capture user impression of robots.}
        The Creepy - Nice scale is from the IEEE Robots database~\cite{IEEERobots}. We added the Boring - Interesting rating based on internal discussions. 
        We also included three ratings to capture users' emotion(s) if the users are touched by the robot~\cite{bradley1994SAM}.  
        Past studies have used custom statements to assess user comfort in physical interactions with robots~\cite{zamani2020effects}. Thus, we added four ratings to capture user-anticipated comfort to touch the robot hand, be touched by the hand, pass or receive objects from the hand (\ie handover), and be present near the robot hand (\ie nearby). 
        Similar to Kalegina \etal~\cite{kalegina2018}, we also asked respondents to provide a descriptive name for the hand and to indicate suitable jobs for it. We do not analyze the provided names and applications in this paper. 
        Table~\ref{tab:questionnaire} presents all the questions and their literature references. 
        We denote the shorthand that correspond to the 17 rating scales with capitalization (\eg Humanlike).

        The questionnaire displayed the hands from one of the eight sets in a random order. 
        Each page showed the edited image of a robot hand at the top and asked the participants to indicate if there is a robot hand and/or object is in the image. Next, the participant answered the questions in Table~\ref{tab:questionnaire} for that hand. As an attention test, we added an extra rating for two of the robot hands in the questionnaire and asked the participants to set its value to ``very uncomfortable (0)''. We also included a dummy blue image instead of a robot hand as an attention test.
        
          \vspace{0.1cm}
       \noindent\textbf{Demographic Questionnaire} -- 
        The demographic questionnaire asked about the participants' age, gender, and the country where they grew up. We also asked them to rate their familiarity with robots on the following scale: (1) None: I have no experience with robots. 
        (2) Novice: I have seen some commercial robots. (3) Beginner: I have interacted with some commercial robots. (4) Intermediate: I have done some designing, building, and/or programming of robots. (5) Expert: I frequently design, build, and/or program robots. Finally, we used the Negative Attitude Toward Robots Scale (NARS) to capture variations between users in their beliefs and feelings toward robots~\cite{RAS2008}. NARS has three subscales capturing negative attitudes toward interacting with robots (S1), social influence of robots (S2), and emotional communication with robots (S3). 
        We included the first subscale from NARS (S1), as it was the most relevant for the evaluation of robot hands.
 
        \subsection{Running an Online Study}  
        We administered the survey online through Amazon Mechanical Turk. The criteria for eligible turkers were having more than 5000 approved hits and a hit rate of 99\% or more. The participants needed to confirm that they are 18 years or older, have normal or corrected to normal vision, and understand English at least at the B2 level. We recruited a total of 168 participants. We removed 8 participants who did not pass our attention tests, resulting in 20 responses for each of the eight sets. 
   
      The majority of the participants were from the United States (122), followed by India (15), Brazil (13), Italy (5), Canada (2), Australia (1), England (1), and Turkey (1). The participants self-identified as man ($n=66$), woman (93), or nonbinary (1). The majority rated their familiarity with robots as novice (65) or beginner (57), followed by no familiarity (23), intermediate (11), or expert (4). The NARS scores were measured on a scale from 6 (the lowest) to 30 (the highest) for a negative attitude toward robots. The mean of the participant scores was 11.75 ($std = 4.88$), indicating positive to neutral attitudes toward robots.

    \begin{figure*}[t!]
    \centering
    \includegraphics[width=\textwidth]{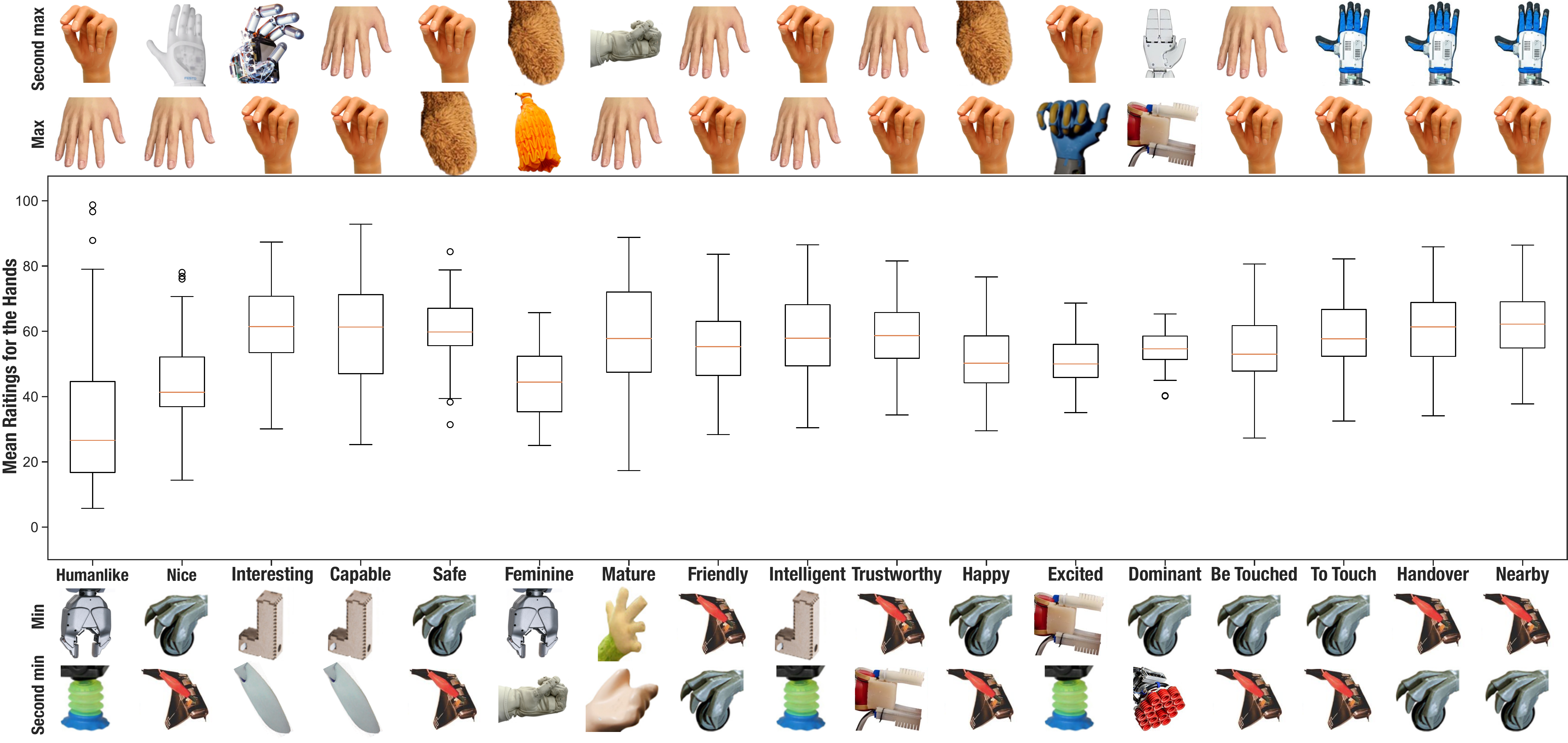}
    \caption{Distribution of the mean user ratings for the 73 hands in the online study. We show the hand images with the highest (max), second to the highest (second max), lowest (min), and second to the lowest (second min) values on the rating scales. }
    \label{fig:summarystats}
    \end{figure*}

\section{Creating a Database and Predictive Models}

We first present the dataset of robot hands and then report our regression models for predicting the user ratings.
\subsection{Database of Design Features and User Ratings}
\label{sec:robothandsdatabase}

We created a dataset with the 73 hands with their ratings and coded the design features of the hands. 
For each hand, we included the average user ratings on the 17 semantic differential scales as well as a link to a reference publication or website. 
Three authors initially agreed on a coding scheme for the hands. Two authors individually coded a random 20\% subset of the hands ($n=16$). They met and discussed the disagreements, clarified the definitions, and merged or divided the features. Next, the same two authors coded another 15\% of the hands ($n=11$). The inter-coder agreement score was 92\%. One of the authors coded the rest of the hands. These authors did not have access to the ratings collected during the online study prior to coding the hands. 
    
This process led to 15 design features for each hand (Table~\ref{tab:handfeatures}). Our focus was on features that can be discerned by a layperson rather than the technical specifications of the hands. Thus, 11 features refer to the visual appearance of the hand, two describe the materials, and one refers to its grasping functionality. We also included whether the hand is a commercial product or not (\eg research prototype). The dataset is available at {\url{http://tiny.cc/RobotHands}. 
 
  \subsection{Constructing Predictive Models}
We developed 17 multiple linear regression models (one for each rating scale) to predict the user ratings for the 73 hands from their design features. 
First, we converted the categorical features into binary representations using a one-hot-encoding scheme. 
Next, we identified the top design features for each rating scale by applying forward and backward stepwise regression on the 1460 (20 users $\times$ 73 hands) ratings in our dataset. The results of this feature selection step provided the design features that could help predict each of the user ratings. 
Finally, we trained a regression model for each of the 17 user ratings.

\begin{figure*}[t!]
\centering
  \includegraphics[width=0.75\textwidth]{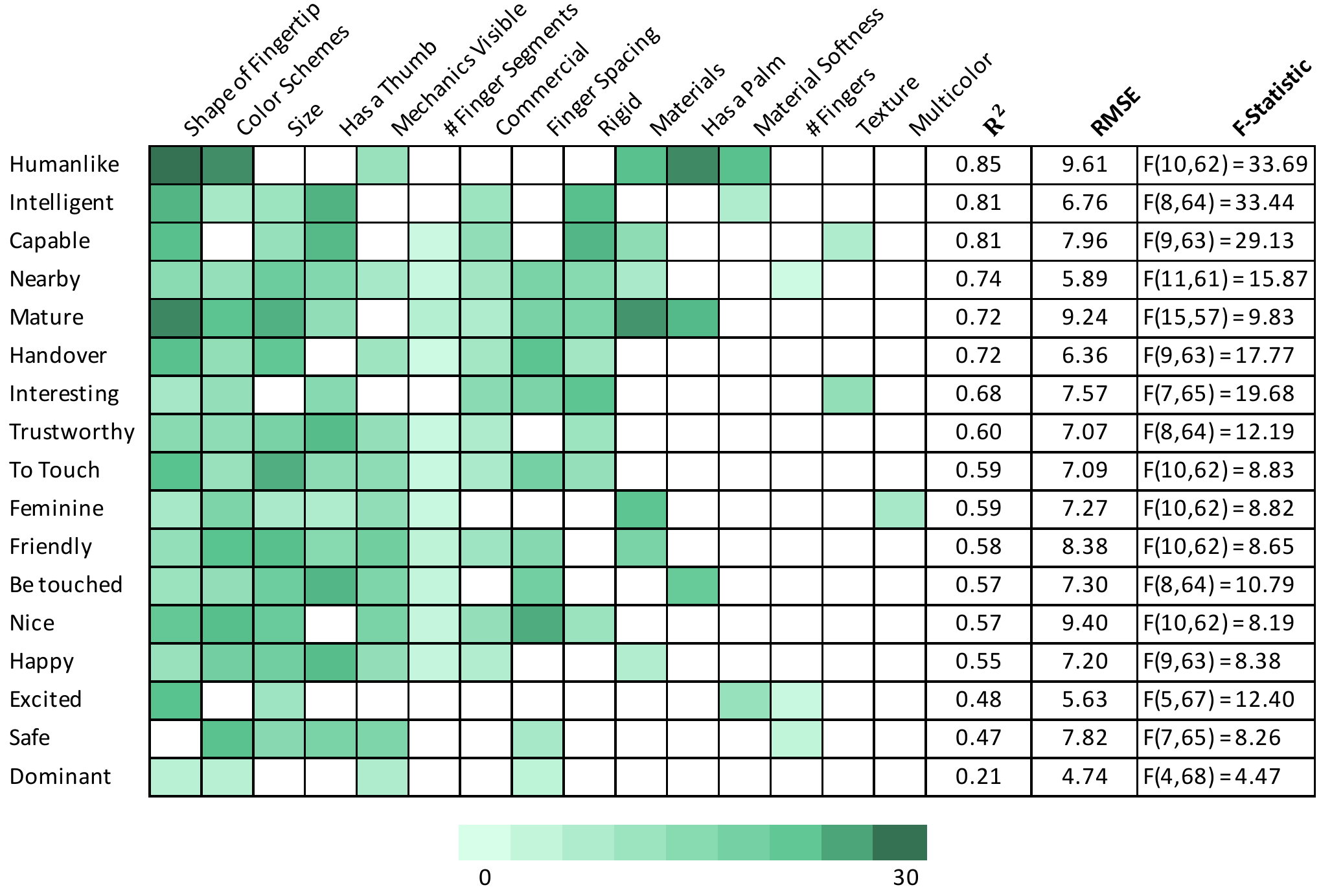}
  \vspace{0.3cm}
  \caption{Contributions of the design features (columns) to the 17 regression models (rows) and the $R^2$, root mean square error (RMSE), and F-statistic for each model. All the regression models are statistically significant ($p< 0.002$). 
  Higher saturation denotes a higher feature weight (0--30). Feature weights are the sum of the absolute values of the feature coefficients in the regression models (\eg The absolute values of the coefficients for Baby, Kid, Adult, and Large are summed to present the contribution of the Size feature to the models). The rows and columns are ordered from high to low values of $R^2$ and feature contributions, respectively.
    }
  \label{tab:heatmap-att-features}
 \end{figure*}

\section{Results}
Figure~\ref{fig:summarystats} presents the distribution of the mean user ratings for the 73 robot hands with example hands from the dataset that fall on the extremes of each rating scale. To inform hand design, we first summarize the performance of our regression models in predicting the user ratings and then present the contribution of each of the design features to the models.

\subsection{Performance of the Regression Models}

Figure~\ref{tab:heatmap-att-features} presents the $R^2$ values, root mean squared error (RMSE), and selected design features for predicting the user ratings on each of the rating scales. 
Over 81\% of the variance among participants' impressions on Humanlike, Intelligent, and Capable ratings is explained by the design features of the hands ($R^2\geq0.81$). 
Also, over 68\% of the variance in the Humanlike, Intelligent, and Capable ratings is accounted for by the design features ($R^2\geq0.68$). 
For seven other rating scales, over 50\% of the variance in the user ratings is explained by the design features of the hands ($R^2>0.5$). 
Finally, less than 50\% of the variance in the Excited, Safe, and Dominant ratings can be explained by the design features ($R^2<0.5$). 
For all the hands, the RMSE over the 17 ratings is below 10 on a 0--100 rating scale. This error is smaller than the standard deviation of the user ratings for each hand in the online study ($std=22.85$).

\subsection{Contribution of the Design Features}
We structure this section around the design features (instead of the user ratings), so that designers can check the effect of a single design feature on the user ratings. We order the design features based on how many user ratings they help predict. For each design feature, we denote this number after the feature name\footnote{The design features for each of the 17 models and their coefficients are included in our dataset.}.

\vspace{0.1cm}
\noindent \textbf{Shape of fingertip (16 models):} The shape of finger/gripper tip helps predict the hand score for 16 (out of 17) rating scales. A round or pointy tip for the fingers is positively associated with the Humanlike, Capable, and Intelligent ratings. A round fingertip also increases user Comfort to Handover objects. Hands with a square fingertip have lower scores on the Interesting, Feminine, and Mature scales, and the users tend to be less Excited and more Dominant if touched by these hands. Other fingertip shapes have lower values for Humanlike, Nice, Mature, Friendly, and Trustworthy ratings. Also, the users are less Happy or Comfortable to Be Touched, Do Handover, or Be Nearby them. When the shape of fingertip is not applicable (\ie the hand does not have any fingers such as a suction cup), the user ratings for Mature, Excited, and Comfortable to Touch the hand decrease. Only the predictions for the Safe scale did not depend on the fingertip shape.

\vspace{0.1cm}
\noindent \textbf{Color scheme (15 models):} The range of skin tone colors, coded as brown in our database, are positively linked to the Humanlike, Nice, Interesting, and Safe ratings. A mix of cool and warm colors lower the Humanlike and Mature scores and increase the Friendly, Happy, and Dominant ratings. Black and gray colors lower the Feminine rating, warm colors (but not skin tone) decrease the Safe rating, while cool colors increase the Intelligent score. Interestingly, the white color has a negative impact on the perception of how Nice, Friendly, and Trustworthy the hand is and how Happy and Comfortable people feel to have physical interactions (To Touch, Be Touched, Handover, Nearby) with it.

\vspace{0.1cm}
\noindent \textbf{Size (14 models):} 
The adult hand size increases the Capable, Mature, and Intelligent scores, and it lowers the Feminine and Comfortable to Touch scores. The kid size is perceived as less Mature, and the users feel less Excited after being touched by these hands. The baby hand size is perceived less Nice, Safe, Friendly, and Trustworthy than the hands with other sizes, and people feel less Happy and Comfortable to have physical interactions with it. The baby hand sizes in our database either had a rigid design or were among the soft manipulators with nuanced working mechanisms. Both designs were negatively evaluated by the users.

\vspace{0.1cm}
\noindent \textbf{Mechanics visible (12 models):} Visible wires and motors make the hand less Humanlike, Nice, Safe, Feminine, Friendly, and Trustworthy. Also, the users feel less Happy, Dominant, and Comfortable to have  physical interactions with the hand compared to when no mechanics are visible.

\vspace{0.1cm}
\noindent \textbf{Has a thumb (12 models):} Hands with a thumb are perceived as more Interesting, Capable, Safe, Feminine, Mature, Friendly, Intelligent, and Trustworthy than those without a thumb. Also, having a thumb improves how Happy and Comfortable the users are to Touch, Be Touched, or Be Nearby the hand.

\vspace{0.1cm}
\noindent \textbf{Number of finger segments (11 models):} Hands with more finger segments are rated as more Capable and Mature but also less Nice, Feminine, Friendly, and Trustworthy. The users are less Happy and less Comfortable to have physical interactions with the hand (To Touch, Be Touched, Handover, Nearby) compared to the hands with fewer segments.

\vspace{0.1cm}
\noindent \textbf{Commercial product (11 models):} In our models, the commercial hands scored higher than the research prototypes on the Nice, Interesting, Capable, Mature, Friendly, Intelligent, Trustworthy, Happy, Comfortable to Touch, Handover, and Nearby user ratings.

\vspace{0.1cm}
\noindent \textbf{Finger spacing (10 models):} 
Uneven spacing of fingers (\eg human hand) is positively linked to the Nice, Mature, and Friendly ratings. The uneven spacing also increases scores on feeling Dominant and Comfortable to have physical interactions with the robotic hand. Even finger spacing (\eg in grippers) lowers the Nice rating. When this features is not applicable (\ie the hand does not have any fingers), the hand is perceived as less Interesting and Safe to users.

\vspace{0.1cm}
\noindent \textbf{Rigid (9 models):} Hands that cannot close or move their fingers score low on Nice, Interesting, Capable, Mature, Intelligent, and Trustworthy ratings. The Comfort to Touch, Handover, or Nearby ratings are lower for these hands too.

\vspace{0.1cm}
\noindent \textbf{Materials (7 models):} Hands with metal components get higher scores on Capable and Mature rating scales and lower scores on the Humanlike, Feminine, Friendly, and Happy ratings compared to hands with other materials. Rubber hands get high scores on Humanlike and Mature, while plastic hands get lower Mature ratings. Hands with other materials get lower scores on the Feminine and Comfortable to be Nearby ratings.

\vspace{0.1cm}
\noindent \textbf{Has a palm (3 models), material softness (3 models), and number of fingers (3 models):} Having a palm has a high weight on the Humanlike score (\ie its coefficient is 24.5 out of 30).
However, having a palm lowers the Mature and Comfortable to Be Touched by the hand ratings. Soft materials increase the Humanlike score, while hard materials increase the Intelligent and Excited ratings. Hands with more fingers get higher scores on the Excited rating and a lower score on the Safe and Comfortable to be Nearby ratings. Surprisingly, the other models do not rely on the number of fingers in their predictions.

\vspace{0.1cm}
\noindent \textbf{Visible texture (2 models) and multicolor (1 model):} Hands without a visible surface texture get lower values on the Interesting and Capable ratings compared to those with a visible texture. Hands without a Multicolor feature get lower Feminine ratings compared to the hands with this feature.

\section{Discussion and Future Work}

We summarize how our findings and predictive models can inform the design of robot hands, discuss the limitations of our work, and outline directions for future studies.

  \subsection{Implications for Robot Hand Design}
 
   The design features selected for the regression models provide guidelines for predicting user ratings of robot hands. Specifically, feature values similar to the human hand create positive impressions (\ie higher values on the rating scales). For example, the round fingertip, brown color, adult hand size, having a thumb, and uneven finger spacing increase the scores for the majority ($\geq11$) of the user ratings. A notable exception is the number of fingers, which does not have much influence on the user ratings. Thus, designers can use fewer or more fingers based on the technical considerations of the target domain. Also, having a palm increases the score for Humanlike and reduces Mature and Comfort to Be Touched ratings, but it has little effect on the other ratings. More finger segments are also associated with higher Capable and Mature ratings, and adding visible surface texture improves Interesting and Capable impressions. On the other hand, some features create negative impressions (\ie lower user ratings) and should be avoided. For example, the white color is frequently used in designing social robots, but our data suggest it is perceived negatively by users. Fingertips that are not round or pointy can also create negative impressions.
    
   Some of the hand features, such as color scheme and surface texture, can be easily modified by designers. Others, including the shape of fingertips, size, surface materials, and visibility of the hand mechanics, can be optimized for both user experience and functionality if these features are considered at the design stage. For example, a round fingertip is linked to positive user impressions and can also have good dexterity and sensing~\cite{fishel2012sensing}. Finally, design features such as if the hand is rigid, if it has a palm, or the number of segments in each finger can fundamentally change the technical specifications. Thus, these features pose a trade-off for design.
    
     These guidelines are based on the visual impression of robot hands. Understanding the visual impression is relevant and important as observing a robot often precedes physical interactions with it. Based on this initial impression, the user may adjust their distance and their physical contact with the hand. This choice also enabled us to collect user input on a wide range of existing robot hands. An interesting question is whether and how the user impressions of the robot hands change after physical contact. A number of haptic and HRI studies suggest that users can infer physical and emotional properties of contact from vision~\cite{tiest2007haptic,vardar2019fingertip,schneider2016hapturk,willemse2016observing}. 
     In contrast, some HRI studies reported that the users rate robots more positively if they are touched by the robot~\cite{shiomi2017effort}. 
     We are currently running an in-lab study to further investigate the impact of physical contact on user ratings.

   \subsection{Limitations and Future Work}

    Our work has three main limitations. First, we did not investigate the effect of grasping and motion parameters on user perceptions of the hands. This was a pragmatic choice. The existing videos of the 73 robot hands in the user study had different viewpoints and demonstrated different object interactions. These differences could unfairly bias user evaluation. Thus, we decided to use images and edit them for consistency in presentation. Second, the 17 rating scales in our study may have limited the participant responses to predefined qualities. We opted for a structured questionnaire to be able to chart user impressions of a large variety of robot hands. A complementary approach would be to conduct open-ended interviews with a smaller subset of representative hands. 
    Third, our results mostly reflect the perception of American users (${>}75\%$ of the participants) who opted to participate in our study. Future studies can investigate whether the same trends hold for other cultures and for those with a negative attitude toward robots.
    
    Additionally, an interesting avenue for future work is user perception of prosthetic hands. 
    Our initial observations suggest that existing prosthetic hand designs are a subset of the robot hands in our database but with more variation in colors and graphical patterns. For example, prosthetic users may have passive limbs, grippers, five-fingered designs, or activity-specific tools as a hand. In our study, we explicitly asked users to imagine that the hands in our database belong to a robot rather than a human. Future work can investigate how an observer's impression of the hand and touch interactions can change when a robotic hand is embodied or controlled by a human rather than a robot. 
  
  \section{Conclusion}
\label{sec:conclusion}
We chart visual impressions of existing robot hands in an online user study. 
These impressions can be estimated from the design features of the hands. 
Our results provide practical design guidelines that can improve the user experience of robots without significantly changing their technical complexity. We share our dataset and predictive models to facilitate further research in this area. 








\balance
\bibliographystyle{IEEEtran}  
\bibliography{references}  

\end{document}